\documentclass[10pt,twocolumn,letterpaper]{article}

\usepackage{times}
\usepackage{epsfig}
\usepackage{graphicx}
\usepackage{amsmath}
\usepackage{amssymb}
\usepackage{tabu}
\usepackage{blindtext}
\usepackage{graphics}
\usepackage{graphicx}
\usepackage{amssymb}
\usepackage{verbatim}     
\usepackage{amsxtra}
\usepackage{epsfig}
\usepackage{amsmath}
\usepackage{latexsym}
\usepackage{ifthen}
\usepackage{psfrag}
\usepackage{color}
\usepackage{xcolor}
\usepackage{array}
\usepackage{blindtext}
\usepackage{multirow}
\usepackage{bm}
\usepackage{nopageno}
\usepackage[linesnumbered,ruled,vlined]{algorithm2e}
\usepackage{array}
\usepackage{listings}
\usepackage{subfig}
\usepackage{fancyhdr}
\usepackage{cite}
\SetCommentSty{mycommfont}

\usepackage{float}

\begin{document}
\title{DP-CGAN : Differentially Private Synthetic Data and Label Generation}
\author{
Reihaneh Torkzadehmahani\\
University of California, Santa Cruz\\
{\tt\small rtorkzad@ucsc.edu}
\and
Peter Kairouz\\
Google AI\\
{\tt\small kairouz@google.com}
\and
Benedict Paten\\
University of California, Santa Cruz\\
{\tt\small bpaten@ucsc.edu}
}

\maketitle
\thispagestyle{fancy}
\begin{abstract}
    Generative Adversarial Networks (GANs) are one of the well-known models to generate synthetic data including images, especially for research communities that cannot use original sensitive datasets because they are not publicly accessible. One of the main challenges in this area is to preserve the privacy of individuals who participate in the training of the GAN models. To address this challenge, we introduce a Differentially Private Conditional GAN (DP-CGAN) training framework based on a new clipping and perturbation strategy, which improves the performance of the model while preserving privacy of the training dataset. DP-CGAN generates both synthetic data and corresponding labels and leverages the recently introduced R\'enyi differential privacy accountant to track the spent privacy budget. The experimental results show that DP-CGAN can generate visually and empirically promising results on the MNIST dataset with a single-digit epsilon parameter in differential privacy.
 
\end{abstract}

\section{Introduction}
    Recent studies have shown that deep neural networks (DNNs) can achieve state-of-the-art performance in various applications such as image recognition\cite{IR01, IR02}, natural language processing\cite{NLP01}, speech recognition\cite{SR01, SR02} and complex video games\cite{GP01, GP02}. It has not only achieved exceptional accuracy in different tasks but also surpassed human-level performance in some of them\cite{GP01, reLU01}. DNNs have also been leveraged in health-related studies ranging from medical images\cite{medical01, medical02, medical03, medical04, medical05} to human genome analyses\cite{gen01, gen02, gen03}.  
    
    Generative Adversarial Networks (GANs)\cite{gan} form a well-researched class of generative models\cite{Vision1, Vision2, Vision3, Vision4}. They can learn the distribution of the training data and generate synthetic data with a distribution very similar to the distribution of the training data. GAN models are particularly used by research communities to generate the synthetic datasets in cases where they cannot directly access sensitive datasets. However, using sensitive data to train GAN models raises privacy concerns for participating individuals. Indeed, recent works show that most machine learning models, including GAN models, are vulnerable to a slew of attacks (from model inversion attacks to membership inference attacks) that can expose significant information about training data\cite{shokri01, LoGAN, Fredrikson, Zhang}. 

    Differential Privacy (DP) \cite{Dwork1, Dwork2} is a common technique to protect the privacy of ML models trained on sensitive data. However, in spite of its popularity, there have been very few recent studies on training GANs in a differentially private way\cite{dpgan, dpgan2019, pate_gan, clinical, BoLi}. The standard procedure leveraged by these recent studies to enforce DP is to first clip the l2 norm of the gradients of the sum of the discriminator's loss on real and fake data and then add Gaussian noise to the clipped gradients.  To keep track of the privacy budget, they typically use the Moment Accountant (MA) technique\cite{abadi}. One of the limitations of these recent works is that they focus exclusively on generating synthetic data (e.g., images) without corresponding labels -- an aspect that renders the synthetically generated data useless for supervised learning applications. More importantly, training high quality GANs with a single digit epsilon parameter (for differential privacy) has been absent so far even for the simplest of all tasks: generating MNIST-like digits.

    In this work, we propose a Differentially Private Conditional GAN (DP-CGAN) training framework, which can preserve the privacy of conditional GAN models using DP\cite{Dwork1, Dwork2}. The main idea in DP-CGAN is that it clips the gradients of discriminator loss on real and fake data separately, which allows the designer to better control the sensitivity of the model to real (sensitive) data. Moreover, DP-CGAN can generate not only synthetic data but also corresponding labels. Further, DP-CGAN employs the newly introduced R\'enyi Differential Privacy (RDP) Accountant\cite{ren} to track the privacy budget. In comparison to the classical MA technique, RDP accounting provides a tighter bound on the privacy budget, allowing for the addition of less noise without compromising the privacy guarantees.
  
    DP-CGAN framework has three main components: conditional generator network, differentially private discriminator network, and privacy accountant. At each step of the training process, the discriminator network is trained in a differentially private manner in which the gradients of loss on real and fake data are clipped separately. Afterwards, the sum of these two set of clipped gradients are computed and noised by adding Gaussian noise to them. Then, the privacy accountant, which is based on the RDP accountant\cite{ren}, is updated by accumulating the spent privacy budget at each step. Next, the generator network is trained with a non-private optimizer. At any given point in time, if the privacy budget exceeds the target one, the training process is halted and the conditional generator network is ready for the creation of synthetic data and labels.
    
    We make the following contributions in this work: 
    \begin{itemize}
      \item We propose DP-CGAN based on a new gradient clipping and noising procedure, which improves the performance compared to the standard procedure to preserve privacy. To the best of our knowledge, DP-CGAN is the first differentially private GAN framework than can generate both the synthetic data and corresponding labels with promising results. It leverages the recently introduced RDP accountant and TensorFlow Privacy\footnote{https://github.com/tensorflow/privacy} package (by Google) to keep track of the privacy budget. 
      
      \item We provide preliminary experimental results showing that DP-CGAN can generate good visual and empirical results on MNIST dataset with single-digit epsilon parameter. This suggests that our work can be viewed as the first stepping stone towards training high quality GANs with strong DP guarantees. 
      
     \item We use the differnetially private conditional generative model to create synthetic data and labels which are used (together) in the training of machine learning models. We test the accuracy of the learned models on real data and show that they perform well. We get an area under the ROC (AUROC) of $87.57\%$ using DP-CGANs compared to $92.17\%$ if we were to train the classifier directly on real data.

    \end{itemize}
    
    The remainder of the paper is organized as follows: Section \ref{sec:Pre} provides a background on GAN, CGAN, and  differential privacy. Section \ref{sec:Rel} overviews the previous related work in the area of preserving the privacy of deep learning models. Section \ref{sec:Our} describes the DP-CGAN framework in detail. Section \ref{sec:Exp} provides the experimental results and Section \ref{sec:Con} concludes the paper with a brief conclusion.

\section{Preliminaries}\label{sec:Pre} 
In this section, we review Generative Adversarial Networks(GAN), Conditional Generative Adversarial Networks(CGAN) and differential privacy concepts used in DP-CGAN.
\subsection{GAN and CGAN}

    Nowadays, there is a great interest in using generative models to create synthetic data that looks like the original one. Generative Adversarial Network(GAN) proposed by Goodfellow et. al~\cite{gan} is one the primary methods to learn generative models for images. GANs consist of two main components: a generator and a discriminator. The generator takes noise as input and generates synthetic data by capturing the original data distribution while the discriminator takes the synthetic data (generator's output) as well as original data (training set) and learns to discriminate between the real (training) and fake (synthetic) data distribution. The discriminator returns two possible values as output which is the assigned score to a test sample representing whether it is real or fake data. The generator and discriminator always try hard to be as accurate as possible and the more the generator improves the quality of the fake data, it gets harder for discriminator to distinguish the difference between the original and fake data. These two components always play a game and are trained simultaneously.
    
    Suppose $p_{z}(z)$ is the probability distribution that random noise $z$ is taken from, $G(z)$ is the generator network that takes the random noise $z$ as input and $D(x)$ is the discriminator network that takes the generator's output as well as the input data $x$ taken form the distribution $p_{data}(x)$. The game that the generator and discriminator play to achieve a trade-off, encapsulates in the following objective function, $V(D, G)$, of a minimax game:
   \begin{equation}
   \begin{split}
        \min_{G}\max_{D}V(D,G)= \\
        \mathop{\displaystyle\mathbb{E}}_{x\sim p_{data}(x)}[log (D(x))] + \\ \mathop{\mathbb{E}}_{z \sim p_{z}(z)}[log(1-D(G(z)))] 
   \end{split}
    \end{equation} 
    
    Conditional GAN\cite{cgan} is an extension of GAN in which both generator and discriminator are conditioned on some side information, “y” that can be any kind of extra information like class labels or data from other modalities. The objective function of a minimax game for CGAN is as the following: \\\\\\
    
    \begin{equation}
    \begin{split}
    \min_{G}\max_{D}V(D,G)=  \\                 
    \mathop{\displaystyle\mathbb{E}}_{x\sim p_{data}(x)}[log (D(x|y))]+ \\
    \mathop{\mathbb{E}}_{z \sim p_{z}(z)}[log(1-D(G(z|y))))] 
    \end{split}
    \end{equation} 
    
\subsection{Differential Privacy}

    Differential privacy\cite{Dwork1, Dwork2} is a mathematical framework to express the level of privacy preservation of individuals in a statistical databases. It provides strong privacy guarantees for algorithms on aggregate databases. Intuitively, in differential privacy, the user should learn about population as a whole but not about particular individual. In other words, if we replace individual $I$ with another random member of the population, the user should learn the same thing about the dataset in presence or absence of individual $I$. Differential privacy has become an actual standard in data protection in both academia and industry\cite{myers:gltr16} (Apple\cite{Apple}, Google\cite{Google} and US Census\cite{USCensus}).\\
    
    \textbf{Definition 1.} (\textit{differential privacy}) A randomized mechanism $M$ over a set of databases $D$, satisfies $(\epsilon,\delta)$-differential privacy if for any two adjacent databases $d , d^{'} \in D$, with only one different sample, and for any subset of output $S \in R$, the following inequality holds:
    
    \begin{equation*}\tag{3}
    Pr[M(d) \in S] \le e^{\epsilon}Pr[M(d') \in S]+\delta
    \end{equation*} 
    
    In pure differential privacy, $\delta =0$ and the additive term $\delta$ does not exist while in approximate differential privacy~\cite{Dwork1}, $\delta$ is used for approximation in the cases that pure differential privacy is broken. $\delta$ is the probability that privacy loss is not bounded by $\epsilon$ and its optimal value is smaller than $\frac{1}{|d|}$ (inverse of the database size).
    
    Differential privacy is resistant to post-processing. That is, any arbitrary randomized mapping of an $(\epsilon, \delta)$-differentially private algorithm, is differentially private as well. 
    
    \textbf{Theorem 1.} (\textit{post-processing}) Given a randomized algorithm $M : D \xrightarrow{}R$ that is $(\epsilon, \delta)$-differentially private and an arbitrary randomized mapping $f : R \xrightarrow{}R^{'}$, $f \circ M : D \xrightarrow{} R^{'}$ is $(\epsilon, \delta)$-differentially private.
    
    A routine approach to privatizing the output of a real-valued function $f: D \xrightarrow{} \mathbb R$ is to add noise with variance in the scale of  $f$'s \textit{sensitivity},  $S_f$, to the output. The sensitivity of a function $f$ is defined as the maximum absolute distance $|f(d) - f(d^{'})|$ ($d$ and $d^{'}$ are adjacent databases). In formal notion:
    \begin{equation*}\tag{4}
     S_f \equiv \max_{d \sim d^{'}} {|f(d) - f(d^{'})|}, 
    \end{equation*} 
    Gaussian noise is one of the popular kinds of noise employed in differential privacy, in which $f(d)$ is perturbed by Gaussian noise $N(0 , {S_f}^2 . \sigma^2)$. That is:
     \begin{equation*}\tag{5}
     M(d) \equiv f(d) + N(0 , {S_f}^2 . \sigma^2)
    \end{equation*} 
    
    Composability is one of the interesting properties of differetnial privacy that makes it possible to combine multiple differentially private mechanisms into one. A standard analysis implies the composition of $k$ mechanisms that each of them are $(\epsilon, \delta)$-differentially private, is at least $(k\epsilon, k\delta)$-differentially private\cite{Dwork1, Dwork2, Dwork4}. One of the possible ways of accounting differential privacy in composition of additive-noise mechanisms is to use Moment Account technique introduced by Abadi et. al\cite{abadi}, which provides strong estimates of privacy loss compared to various versions of composition theorem~\cite{Dwork1, Dwork4, peter_composition, Dwork5, strong_composition} including strong composition theorem \cite{strong_composition}. RDP accountant \cite{ren} is a new approach based on a new definition of privacy,  R\'enyi differential privacy, which  provides a tighter bound for privacy loss in comparison with Moment Accountant. 

\section{Related Work}\label{sec:Rel} 

    Some previous studies have proposed approaches to addressing the problem of preserving privacy in Deep Learning. Shokri et al.\cite{shokri01} developed a distributed approach in which multiple parties train a model on their local training set independently. Then, each party selects a set of key parameters, and shares them with the other parties. 
    Although this method has high training accuracy without sharing the input parameters, Abadi et al.\cite{abadi} showed that the overall privacy loss for each party exceeds several thousands on MNIST dataset using Moment Accountant technique they introduced.
    
    Moment accountant mechanism\cite{abadi} can be used to track the overall spent privacy budget, $(\epsilon, \delta)$, for composing Gaussian Mechanisms with random sampling (e.g. training process in Stochastic Gradient Descent). This method provides a much tighter estimation for privacy loss compared to standard composition theorem\cite{composition}.  It computes the log moments of the random variable indicating privacy loss and then calculates the tail bound using moments bound and standard Markov inequality. The result is privacy loss estimation in terms of differential privacy. In addition to Moment Accountant technique, Abadi et al.\cite{abadi} proposed a method to make the Stochastic Gradient Descent(SGD) process differentially private. 
    
    Private Aggregation of Teacher Ensembles (PATE)\cite{pate} is a framework that leverages the moment accountant mechanism to trace the privacy leakage of knowledge transfer task using differential privacy. It presents a differentially private semi-supervised learning method in which the training data is split into multiple disjoint sets and the teacher models are trained independently. The teacher ensemble predicts the labels after perturbing counts of teachers' votes by Laplace noise while the student model is trained on public data as well as labeled data from the teacher model and can be published publicly. Although this method outperforms Shokri et al.\cite{shokri01} work in terms of both accuracy and privacy, it assumes the model has access to public data which may not be the case in practice. Moreover, the teacher ensemble just responds to the queries for which the consensus among teachers is sufficiently high. 
    
    Some other previous researches focused on preserving privacy of GANs in particular. DPGAN method\cite{dpgan}, enforces differential privacy during the training process of the discriminator by adding Gaussian noise to the gradient of Wasserstein distance in WGAN algorithm and uses post-processing theorem to guarantee differential privacy for the generator. However, it is unclear how the overall privacy budget is accounted, the results do not look promising even on MNIST dataset and there is no methodology for creating labels for synthetic images. 
    
    Similar to DPGAN method, PATE-GAN approach\cite{pate_gan}  enforces privacy by making the discriminator differentially private. In PATE-GAN, the discriminator is replaced with modified version of PATE\cite{pate} in which the student model allows back-propagation to the generator and there is no need to have access to public training data. It employs the generated data to train different classifiers and evaluate the quality of generated data by testing these classifiers on real test data. The limitation of PATE-GAN is that it assigns binary labels for synthetic data, and therefore, it is not applicable for multi-label datasets. Moreover, the datasets used to evaluate the model are small.  The other work is a DPGAN framework for time series, continuous, and discrete data\cite{dpgan2019}. This framework is alike the previous DPGAN work\cite{dpgan} except it employs moments accountant approach to account the privacy budget and clips the discriminator gradients while reducing the clipping parameter over time (adaptive clipping). 
        
    Unlike  DPGAN method\cite{dpgan}, our proposed method leverages RDP accountant technique to follow the consumed privacy budget, $(\epsilon, \delta)$ and generates not only synthetic data but also the labels using a Conditional GAN model. In contrast to PATE-GAN\cite{pate_gan} which generates only binary labels, our model generates multi-class labels. Finally, in DPGAN frameworks\cite{dpgan, dpgan2019} the discriminator gradients are clipped and perturbed by adding Gaussian noise to gradients of the discriminator loss, while in our framework, Gaussian noise is added to the accumulation of clipped gradients of discriminator loss on real data and clipped gradients of discriminator loss on fake data. 

\section{Our Approach}\label{sec:Our} 

    As mentioned before, DP-CGAN can generate the synthetic data as well as the corresponding labels while preserve privacy of training samples. To this end, the DP-CGAN makes the training process private by injecting random Gaussian noise into the optimization process of the discriminator network. Based on post-processing theorem\cite{Dwork3} making the generative network differentially private results in having a differentially private generator too.
   DP-CGAN tracks the spent privacy loss using RDP accounting technique\cite{ren}, which provides tighter estimation on privacy loss in comparison with moment accountant technique. The training procedure stops if the spent privacy budget $(\epsilon, \delta)$ goes beyond the target ones. 
    \begin{algorithm}
    \caption{DP-CGAN}
    \DontPrintSemicolon
    Input: Examples $\{x_1 , x_2, ..., x_N\}$, labels $\{y_1, y_2, ..., y_N\}$, target epsilon $\epsilon$, target delta $\delta$, noise scale $\sigma$, clip norm bound  $C$, learning rate $lr$, batch size $bs$
      
    Output: Differentially private Generator that generates synthetic data and labels
      
      should$\_$terminate = False
      
      \While{$(step \le max\_step \And  !\ should\_terminate)$}
      {
        \normalfont{- Sample random batch $(X^t, Y^t)$ of size $bs$ with probability $bs/N$ from data distribution $p_{data}{(X)}$ } \newline
           
        \normalfont{- Sample noise batch $Z_t$ of size $bs$ from noise prior $p_{z}{(z)}$ }
            
        \tcc{Update the Discriminator Network}
            
        $d\_loss\_real \xleftarrow[]{} \log(D(X^{t}))$
            
        $d\_loss\_fake \xleftarrow[]{} \log (1- D(G(Z^{t})))$
            
        \textbf{Compute per-example gradients of discriminator loss on real data $X_t$ and clip them}
            
        \For{$i \in X_t$}
        {
            Compute ${grad_{d\_real}}^{t} \xleftarrow[]{} \nabla_{\theta_d}d\_{loss}\_real ({{\theta_d}^{t}}, X_{i})$
        } 
            
        ${grad_{d\_real}}^t = {grad_{d\_real}}^t / max(1, \frac{||grad_{d\_real}||_{2}}{C}) $            
        \textbf{Compute per-example gradients of discriminator loss on fake data $Z_t$ and clip them }
            
        \For{$i \in Z_t$}
        {
            Compute ${grad_{d\_fake}}^{t} \xleftarrow[]{} \nabla_{\theta_d}d\_{loss}\_fake ({{\theta_d}^{t}}, Z_{i})$
        }

        ${grad_{d\_fake}}^t = {grad_{d\_fake}}^t / max(1, \frac{||grad_{d\_fake}||_{2}}{C}) $
            
        \textbf{Compute the overall gradients of discriminator and add Gaussian Noise to them}

        ${grad_{{d}}}^t \xleftarrow[]{}{\frac{1}{bs}{\sum {grad_{{d\_real}}}^t + {grad_{{d\_fake}}}^t + N (0, \sigma^2 C^2 I)}}$
            
        \textbf{Take the gradient Descent step for discriminator}
            
        ${{\theta_{d}}_{t+1}} \xleftarrow[]{} SGD(grads\_d^t, {{\theta_d}_t}, lr)) $
            
        \tcc{Update RDP Accountant}

        \textbf{Accumulate the spent privacy budget using RDP Accountant}
            
        \tcc{Update the Generator Network}
            
        $g\_loss \xleftarrow[]{} log (1- D(G(Z^{t})))$
             
        \textbf{Compute gradients of generator loss}
        
        Compute ${grad\_g}^{t} \xleftarrow[]{} \nabla_{\theta_g}g\_loss ({{\theta_g}^t}, Z_{i})$
         
        \textbf{Take the gradient Descent step for generator}
            
        ${{\theta_g}^{t+1}} \xleftarrow[]{} ADAM({grad\_g}^t, {\theta_g}^t)$
            
        \If{ $spent\_epsilon > \epsilon$ OR $spent\_delta > \delta$}{
            
            \tcc{Running out of privacy budget}
        
            should$\_$terminate = True
            }}
    \end{algorithm}
    DP-CGAN makes the optimization process of discriminator loss (discriminator training) differentially private by computing the per-example gradients of the discriminator loss on both real and fake data, clipping the per-example gradients on real data and fake data separately, summing up two sets of the clipped gradients, perturbing the clipped gradients by adding Gaussian noise $N(0, \sigma^2 C^2)$, $\sigma$ is noise multiplier and $C$ is clipping value,  to them, and finally applying the perturbed gradients. 
    
    Algorithm 1 outlines the training process of DP-CGAN. According to the algorithm, the model updates the discriminator network and the generator network as long as the number of iterations is less than maximum iteration count and the spent privacy budget is less than the target $\epsilon$. At each step, it minimizes the discriminator loss function by computing the discriminator gradients of loss on real data and clipping them by $L_2$-norm (lines 9-12), computing the discriminator gradients of loss on fake data(lines 13-15) and clipping them by $L_2$-norm, compute the overall clipped gradients of discriminator by adding these two sets of clipped gradients, adding Gaussian noise to them and taking average over all the perturbed clipped per-example gradients in the batch(line 16-17), and finally applying the gradients (line 18). The model tracks the spent privacy budget by accumulating the spent privacy budget and updating the RDP accountant every time that noise is injected into the model(line 20). Then, the generator the gradients of generator loss are computed and applied so that the generator network gets trained(line 21-25). The last step is to check the overall spent privacy budget so far. If the spent $\epsilon$ or the spent $\delta$ has exceeded the target values, training is stopped, otherwise it continues (line 26-27). 

\section{Experimental Results}\label{sec:Exp} 
\begin{figure}[ht]
          \centering
              \includegraphics[width=1.03\columnwidth]{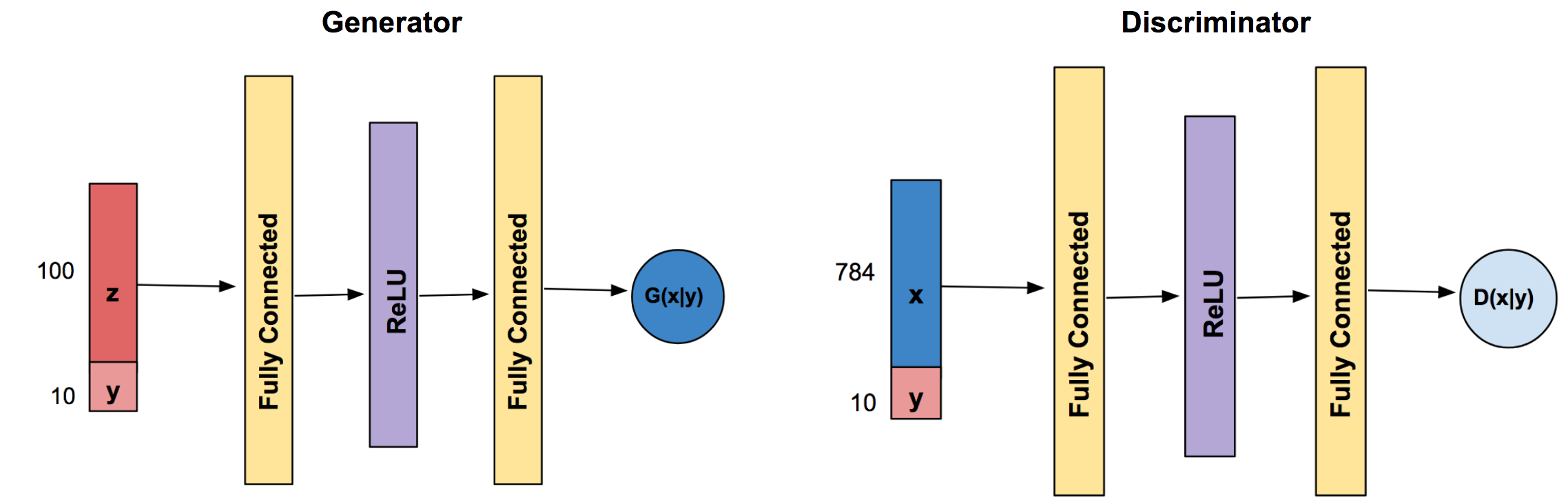}
          \caption{Vanilla CGAN Generator and Discriminator Architecture }
          \label{figure1}
    \end{figure}

    We compare the performance of DP-CGAN to CGAN with no privacy and CGAN trained with standard differentially private approach.The CGAN architecture used in all models is a vanilla CGAN in which both generator and discriminator consist of two fully connected layers.The generator takes random noise sample $z$ and the corresponding label $y$ as inputs while the discriminator inputs are real training sample $x$ and its label $y$. Figure \ref{figure1} depicts the generator and discriminator architecture of the vanilla CGAN. 
    Differentially private CGAN models use the new privacy package of TensorFlow Privacy (by Google), a python library that includes the implementation of few differentially private optimizers as well as the privacy accountants to keep track of the privacy loss. They leverage differentially private Gradient Descent as optimizer and RDP accountant as privacy accountant from this package. 
    
   The dataset used used in the evaluation is MNIST handwritten dataset containing 60k training samples and 10k test samples. In the experiments, batch size is set to 600, $\delta = 10^{-5}$ and learning rate is set by an adapative approach in which the initial learning rate is 0.15, it is decreased to 0.052 in iteration 10K and is fixed on 0.052 for the rest iterations. 
   
   We trained Logistic Regression and Multi-Layer Perceptron classifiers using the synthetic data and labels generated by the models and tested the classifier on real test data. Closer performance of the classifier trained on synthetic data generated by differntially private models and on real data indicates that the model has captured the real data distribution better. We measured the performance of the classifier using the Area under ROC curve metric (AuROC). In the evaluation process, the generative model takes the 60k MNIST training data and the labels as input and generates 60k synthetic labeled data.Then, the classifier is trained on the generated data. Finally, performance of the trained classifier is evaluated on the 10k test data using AuROC metric. 
   
   Table 1. lists the results of AuROC for the three models as well as the case in which classifiers are trained on real data. According to the table, the AuROC of DP-CGAN  is higher than CGAN trained with basic differentially private method, indicating that new clipping and perturbing technique used in DP-CGAN improves the performance. On the other hand, the AuROC of DP-CGAN is about $5\%$ lower than that for real data and this is the price we pay to have privacy. 
   
   \begin{table}[ht!]
    \begin{center}
    \begin{tabular}{|c|c|c|c|c|}
    \hline
      & Real &  CGAN & \textbf{DP-CGAN} & \multicolumn{1}{|p{1.5cm}|}{\centering CGAN with \\ basic DP}\\
    \hline\hline
    LR &  $92.17\%$ & $91.10\%$ & $\bm{87.57\%}$ &  $83.42\%$ \\
    \hline
    MLP & $97.60\%$ & $91.06\%$ & $\bm{88.16\%}$ & $83.29\%$\\
    \hline
    \end{tabular}
    \end{center}
    \caption{Comparing AuROC for Logistic Regression(LR) and Multi-Layer Perceptron(MLP), which are trained on real data, data generated by CGAN (non-private), DP-CGAN and CGAN with basic differentially private approach using $\epsilon = 9.6$ , and $\delta = {10^{-5}}$} 
    \end{table}

    We also visualized the images generated by the models (Figure 2) . In the figure, the most left column shows the results for DP-CGAN, the left column represents the results for CGAN with no privacy, and the right column depicts the synthetic images generated by CGAN with basic differentially private approach. According to the figure,  the quality of the images generated by DP-CGAN is better than CGAN with basic differentially private approach but worse than CGAN with no privacy.  
    \begin{figure}[h!]
  \subfloat[]{\includegraphics[width=0.05\textwidth]{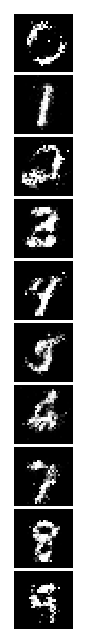}\label{subimg:dp-cgan}}
     \hfill
  \subfloat[]{\includegraphics[width=0.05\textwidth]{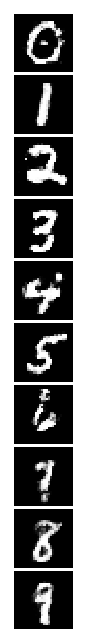}\label{subimg:cgan}}
     \hfill
  \subfloat[]{\includegraphics[width=0.05\textwidth]{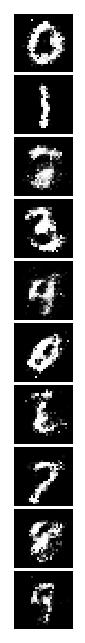}\label{subimg:base-dp}}
     
  \caption{(a) DP-CGAN, (b) CGAN with no privacy, (c) CGAN with basic differentially private approach}
\end{figure}

\section{Conclusion}\label{sec:Con} 
    In this research, we proposed DP-CGAN framework that is a differentially private GAN model capable of generating both synthetic data and corresponding labels. The main idea behind DP-CGAN is that it clips the gradients of discriminator loss on real and fake data separately, sums up two sets of gradients, and adds Gaussian noise to the sum. DP-CGAN employs RDP account technique to track the spent privacy budget. The experimental results showed that DP-CGAN improves the performance compared to basic DP-CGAN and generates promising results on MNIST dataset. 
  
    The architectures we used for the generator and discriminator are rather simple. We are going to consider deep CGAN architectures with multiple convolutional layers to improve the quality of the synthetic data while spending the same privacy budget as we did for vanilla CGAN. Moreover, our results are still preliminary and we are going to show high quality differentially private CGANs on more challenging datasets such as CIFAR100 and CelebA/B. Finally, our preliminary results are very promising and we can extend our methodology to tackle the mentioned challenges.

{\small
\bibliographystyle{ieeetr}
\bibliography{main}
}

\end{document}